\documentclass[sigconf]{acmart}

\acmSubmissionID{735}

\usepackage{booktabs} 

\usepackage{bm} 
\usepackage{mathtools}
\usepackage{amsmath}
\usepackage{multirow}
\usepackage{balance}
\usepackage{enumitem} 
\usepackage{graphics} 
\usepackage{xcolor}  

\def\projname{Ultra Inertial Poser\xspace}

\ifx\DRAFT\fi

\newcommand{\cmt}[4]{\ifx\DRAFT\undefined\else\colorbox{#3}{\textcolor{#4}{\small{\textsf{[\textbf{#1}: #2]}}}}\fi}
\newcommand{\ph}[1]{\ifx\DRAFT\undefined\else\colorbox{purple}{\textcolor{white}{\small{\textsf{#1}}}}\fi}

\usepackage[ruled]{algorithm2e} 

\SetAlFnt{\small}
\SetAlCapFnt{\small}
\SetAlCapNameFnt{\small}
\SetAlCapHSkip{0pt}

\copyrightyear{2024}
\acmYear{2024}
\setcopyright{acmlicensed}\acmConference[SIGGRAPH Conference Papers '24]{Special Interest Group on Computer Graphics and Interactive Techniques Conference Conference Papers '24}{July 27-August 1, 2024}{Denver, CO, USA}
\acmBooktitle{Special Interest Group on Computer Graphics and Interactive Techniques Conference Conference Papers '24 (SIGGRAPH Conference Papers '24), July 27-August 1, 2024, Denver, CO, USA}
\acmDOI{10.1145/3641519.3657465}
\acmISBN{979-8-4007-0525-0/24/07}

\acmDOI{10.1145/3641519.3657465}

\begin{document}
\title[Ultra Inertial Poser: Motion Tracking from Sparse IMU and Ultra-Wideband Ranging]{Ultra Inertial Poser: Scalable Motion Capture and Tracking from Sparse Inertial Sensors and Ultra-Wideband Ranging}

\author{\href{https://orcid.org/0000-0001-6965-4828}{Rayan Armani}, \href{https://orcid.org/0009-0006-4670-2376}{Changlin Qian}, \href{https://orcid.org/0000-0001-6000-4474}{Jiaxi Jiang}, and \href{https://orcid.org/0000-0001-9655-9519}{Christian Holz}}
\affiliation{%
 \institution{Department of Computer Science, ETH Zürich
    \country{Switzerland}}
}


\renewcommand\shortauthors{Armani et al.}
\begin{teaserfigure}
    \vspace{-2mm}
  \includegraphics[width=\textwidth]{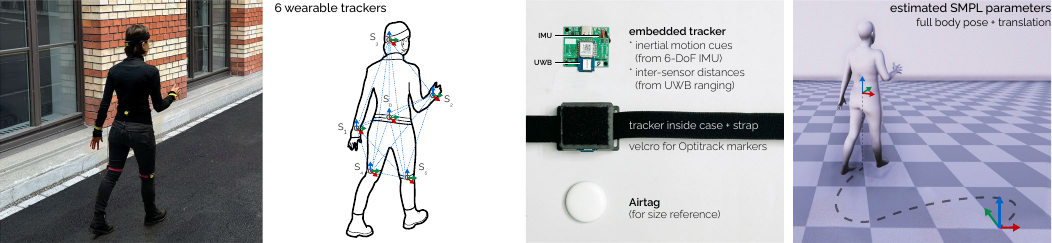}%
  \vspace{-2mm}
  \caption{Our method \emph{Ultra Inertial Poser} estimates 3D full body poses and global translation (i.e., SMPL parameters) from the inertial measurements on a sparse set of wearable sensors, augmented and stabilized by the estimated inter-sensor distances based on UWB ranging.
  Our lightweight standalone sensors stream \emph{raw IMU signals}, from which we estimate each tracker's 3D state and fuse it with acceleration and inter-sensor distances in a graph-based machine learning model for pose estimation.}
  \Description{This is the teaser figure for the article in four sections.
  The left-most image is a photograph of an individual walking outdoors wearing six wearable trackers placed on the head, back, wrists, and knees. The image next to it is a stylized outline of the person, with reference coordinates for each sensor $S_0$ to $S_5$ and lines connecting them illustrating inter-sensor distances. Next, is a detailed look of the sensor hardware, comprising an IMU, UWB module, and a case with a strap, alongside Velcro for attaching Optitrack markers. An AirTag is included for size reference. Finally, the right-most image shows the SMPL model estimated using data from the wearable sensors, depicting full-body pose and translation.
}
  \label{fig:teaser}
\end{teaserfigure}

\begin{abstract}

While camera-based capture systems remain the gold standard for recording human motion, learning-based tracking systems based on sparse wearable sensors are gaining popularity.
Most commonly, they use inertial sensors, whose propensity for drift and jitter have so far limited tracking accuracy.
In this paper, we propose \emph{Ultra Inertial Poser}, a novel 3D full body pose estimation method that constrains drift and jitter in inertial tracking via inter-sensor distances.
We estimate these distances across sparse sensor setups using a lightweight embedded tracker that augments inexpensive off-the-shelf 6D inertial measurement units with ultra-wideband radio-based ranging---dynamically and without the need for stationary reference anchors.
Our method then fuses these inter-sensor distances with the 3D states estimated from each sensor.
Our graph-based machine learning model processes the 3D states and distances to estimate a person's 3D full body pose and translation.
To train our model, we synthesize inertial measurements and distance estimates from the motion capture database AMASS.
For evaluation, we contribute a novel motion dataset of 10 participants who performed 25 motion types, captured by 6 wearable IMU+UWB trackers and an optical motion capture system, totaling 200 minutes of synchronized sensor data (UIP-DB).
Our extensive experiments show state-of-the-art performance for our method over PIP and TIP, reducing position error from 13.62 to 10.65\,cm (22\% better) and lowering jitter from 1.56 to 0.055\,km/s$^3$ (a reduction of 97\%).

\begin{center}
~\\
\vspace{1mm}
UIP code, UIP-DB dataset, and hardware design:

\href{https://github.com/eth-siplab/UltraInertialPoser}{\texttt{\color{magenta}{https://github.com/eth-siplab/UltraInertialPoser}}}
\vspace{3mm}
\end{center}

\end{abstract}

%
%

%
%


\maketitle

\section{Introduction}
\label{sec:introduction}

Accurate and unrestricted motion tracking is fundamental across many domains and applications, such as animation, games, Augmented and Virtual Reality as well as fitness training and rehabilitation.
High-quality motion capture systems rely on cameras, involving a tethered, comprehensive, and stationary setup~\cite{optitrack, vicon}.
Compensating for the worn capture suits or markers these systems typically require, researchers have instead explored markerless motion capture from images~\cite{kanazawa2018end, chen20173d} and videos~\cite{kocabas2020vibe, habermann2019livecap} from one~\cite{habermann2020deepcap, li2020cascaded, hu2021conditional} or more cameras~\cite{de2008performance}.

Beyond stationary setups, tracking systems built around wearable sensors have leveraged miniaturized cameras in head-mounted devices~\cite{10.1145/2980179.2980235, tome2019egopose} or body-worn sensors~\cite{shiratori2011motion,zhang2020fusing,li2020mobile,yi2023EgoLocate}.
Such motion capture is mobile and requires less instrumentation for tracking, though at the cost of substantial motion artifacts and self-occlusion in sensor observations~\cite{wang2021estimating}.

Substituting wearable cameras with even smaller motion sensors, dominantly inertial measurement units (IMU), has enabled less expensive setups but requires more comprehensive full-body sensor coverage, usually attached to motion capture suits to achieve comparable quality (e.g., 17--19 sensors for Xsens~\cite{XSens} or Noitom~\cite{noitom}).
The multitude of IMUs is needed to compensate for the diminished motion cues, which commercial products stabilize with kinematic models to fuse cross-sensor observations.

To lessen the need for comprehensive sensor coverage, recent methods have proposed estimating full-body poses from only a sparse set of inertial sensors with learning-based techniques~\cite{huang2018deep, yi2021transpose, yi2022physical}.
It is worth noting that the quality of their estimates thereby rely on the well-calibrated \emph{absolute 3D state} provided by the IMU sensors, which has dominantly been from Xsens' proprietary units.
Even so, without access to direct position observations and due to the nature of inertial sensing, current pose estimates suffer from overall drift and joint jitter as well as degrading accuracy when body motions are performed slowly.

In this paper, we propose \emph{Ultra Inertial Poser}, a learning-based method for 3D full body pose estimation that fuses \emph{raw} IMU readings with inter-sensor distances for stable predictions.
Our method estimates these distances from ultra-wideband (UWB) ranging, which we add to a small, inexpensive, and untethered embedded sensor node (\autoref{fig:teaser}).
Our graph-based neural model fuses these distances with our 3D state estimations from the raw IMU signals to recover the SMPL parameters of body pose and global translation.

For training, we use AMASS~\cite{AMASS:ICCV:2019} to synthesize IMU signals and distances and augment them with our empirically validated and collision-aware noise model.
For evaluation, we introduce a novel motion capture dataset in which participants performed a long list of activities.
Our method surpasses the current SOTA methods PIP~\cite{yi2022physical} and TIP~\cite{jiang2022transformer} in position accuracy (22\% lower error) and jitter (97\% reduction).

~\\
We make the following contributions in this paper:
\begin{enumerate}[leftmargin=*,nosep]
    \item \emph{Ultra Inertial Poser}, a novel method to incorporate distance constraints into a pose estimation framework for inertial measurements.
    Our graph-based neural network affords effective training using existing large motion capture datasets (e.g., AMASS), complemented by synthetic and noise-augmented distances.
    
    \item An embedded sensing platform with \emph{inexpensive off-the-shelf components} for UWB-based ranging and IMU-based motion detection similar to those inside UWB-based item trackers.
    Our UWB communication protocol and distance estimation requires \emph{no stationary nodes} 
    and dynamically compensates for moments of occlusion between nodes as ranging crosses the human body.
    Compared to electromagnetic sensing, UWB ranging is lower power, less susceptible to interference, and can \emph{scale} simply.
    
    \item \projname's pipeline is first to operate on the \emph{raw signals} from IMU sensors (i.e., \emph{just} acceleration, angular velocity), and UWB radios---without requiring proprietary components to obtain global 3D orientation as input (e.g., Xsens in prior work).
    
    \item \emph{UIP-DB}, a novel motion dataset of 25 types of motion activities from 10 participants, including everyday movements as well as challenging motions for IMU-only tracking (e.g., slow transitions).
    UIP-DB contains synchronized 6-DoF IMU signals, UWB measurements and distances, as well as SMPL references based on a 20-camera Optitrack setup, totaling 200 minutes of data.
\end{enumerate}

\vspace{1mm}
\noindent
Taken together, Ultra Inertial Poser is a scalable approach for inexpensive full-body motion tracking using sparse wearable sensors.
Our tracking approach is wireless and affords motion capture in the wild outside controlled indoor environments (Figure~\ref{fig:video}).

\section{Related Work}
\label{sec:related_work}

\projname\ is related to human body pose estimation using wearable sensors and positioning based on ultra-wideband ranging.
\vspace{-2mm}

\begin{figure*}[t]
    \centering
    \includegraphics[width=\linewidth]{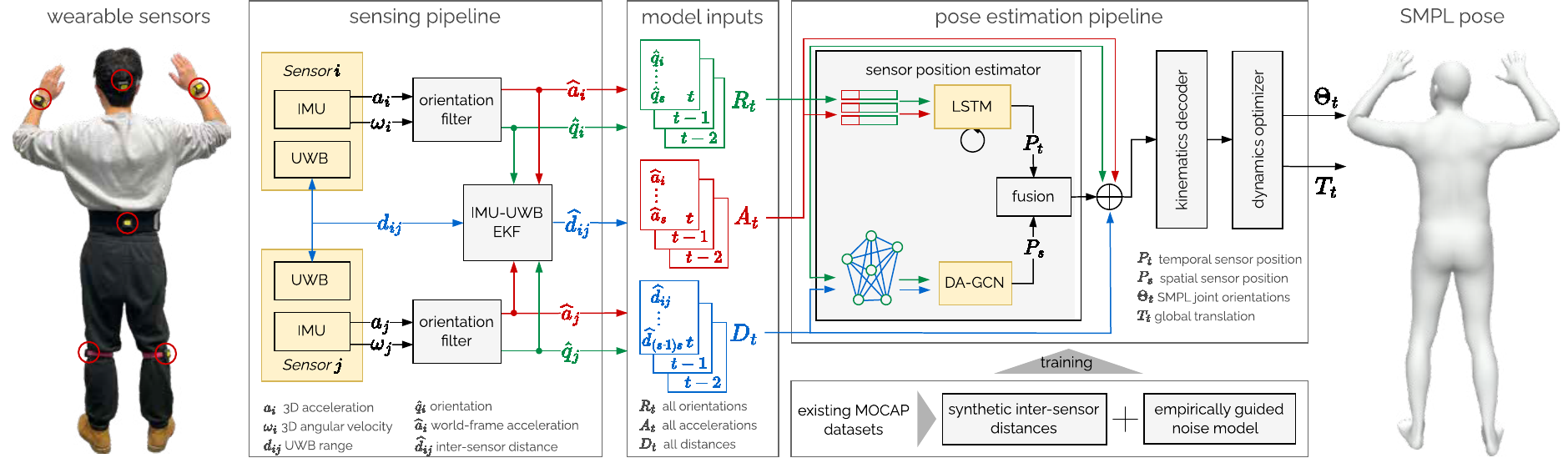}
    \caption{Overview of our method.
        The person is wearing 6 sensor nodes, each featuring an IMU and a UWB radio.
        Our method processes the raw data from each sensor to estimate sequences of global orientations, global accelerations, and inter-sensor distances.
        These serve as input into our learning-based pose estimation to predict leaf-joint angles as well as the global root orientation and translation as SMPL parameters.}
    \label{fig:pipeline}
    \Description{
    The figure is a diagram giving an overview of our multi-stage processing pipeline for pose estimation.
    On the left, a person is shown wearing 6 sensors attached to the head, lower back, wrists, and knees, highlighted with red circles.
    The first stage of the pipeline is the sensing pipeline. The diagram shows details for a pair of sensors $i$ and $j$, each consisting of an IMU and UWB modules. The acceleration and angular velocity of each sensor are fed to a VQF filter to determine each sensor's orientation and world-frame acceleration. The raw UWB ranges between the two sensors is fused with the orientations and world-frame accelerations of each sensor through an IMU-UWB Extended Kalman Filter to obtain refined inter-sensor distance estimates. 
    The orientations, world-frame accelerations, and distance estimates from all 6 sensors are concatenated at each timestep into $R_t$, $A_t$ and $D_t$ arrays respectively to form the inputs for the next stage of the pipeline.
    The second stage of the pipeline is pose estimation pipeline. Its first block is the sensor position estimator which is split into two branches. In one branch,  $R_t$ and $A_t$ are concatenated and passed through an LSTM to give $P_t$, a temporal estimate of sensor positions relative to the pelvis. In the other branch, $R_t$ and $D_t$ form a graph that is passed through a GCN to obtain $P_s$, a spatial estimate of relative sensor positions. $P_s$ and $P_t$ estimates are then fused into a single sensor position estimate. This estimate, and the inputs $R_t$, $A_t$ and $D_t$ are then passed to a kinematics decoder followed by a dynamics optimizer to obtain the SMPL pose parameters and global translation. These final outputs are represented by an avatar rendering the pose of the person at the start of the pipeline. 
    The training process for the pose estimation pipeline, outlined at the bottom of the diagram, leverages existing MOCAP datasets to synthesize IMU data and inter-sensor distances, incorporating an empirically derived noise model.
    }
\end{figure*}

\subsection{Motion Capture from Inertial Sensors}


Inertial Measurement Units (IMUs) are popular for tracking motion due to their compact size, low power consumption, and affordability.
They combine accelerometers and gyroscopes to track changes in acceleration and angular velocity, often supplemented by magnetometers to measure surrounding magnetic fields.
However, IMUs cannot directly observe positions, and position estimates suffer from drift.
Commercial systems address this by integrating data from 17--19 IMUs with biomechanical models (XSens~\cite{XSens}).

Pose estimation from sparse sets of IMUs has become viable using the large available motion capture (mocap) datasets to train learning-based models~\cite{AMASS:ICCV:2019, huang2018deep,trumble2017total, guzov2021human,streli2023hoov,mollyn2023imuposer}.
Early systems used offline optimization to fit the SMPL body model to data from six IMUs~(e.g., SIP~\cite{von2017sparse}).
Deep Inertial Poser's (DIP) bidirectional recurrent neural network maps IMU readings to local joint motions~\cite{huang2018deep}, which other methods further improved~\cite{nagaraj2020rnn}.
More recent methods also use 6 IMUs and optimize both pose and translation estimation via ground contact points ~\cite{yi2021transpose, jiang2022transformer} and physical constraints~\cite{yi2022physical}.
With the advance of VR/AR applications, several methods further sparsified input requirements to just the upper body to estimate full-body pose from head and hand poses~\cite{yang2021lobstr,ahuja2021coolmoves,jiang2022avatarposer,jiang2023egoposer,du2023agrol, zheng2023realistic}.

To combat the persistent challenges in estimating accurate joint angles and positions, researchers have revisited external~\cite{pan2023fusingmono,von2018recovering} or body-worn~\cite{yi2023EgoLocate} cameras for visual-inertial tracking.
Alternatively, EM-Pose~\cite{kaufmann2021pose} measures the relative 3D offsets and orientations between joints using 6--12 custom electromagnetic (EM) field-based sensors.

Given the difficulties of reproducing custom EM sensing technology with commodity components, our work instead builds on previous approaches that estimate mere distances between wearable trackers (e.g., using ultrasonic sensors~\cite{vlasic2007practical, liu2011realtime}).
Our method obtains such pairwise distances from ultra wideband-based ranging and integrates them as constraints into a learning-based sparse inertial sensing pipeline.
The small size, inexpensive nature, and low power consumption of UWB radios can thus complement IMUs as part of wearable and mobile devices.

\subsection{Ultra-Wideband Ranging}

Ultra-wideband (UWB) signals are characterized by their extensive bandwidth (larger than 500\,MHz) and very short duration waveforms, typically in the order of a nanosecond~\cite{uwb_textbook}.
This characteristic allows for accurate determination of the time of departure and arrival of signals, making UWB systems ideal for ranging applications through a variety protocols, including Two-Way Ranging (TWR)\cite{ieee802.15.4a}, Time Difference of Arrival (TDoA)\cite{ubisense,tiemann2017tdoa}, and Angle of Arrival (AoA)~\cite{Ledergerber2019a0a}.
Additionally, UWB radios are compact and have low power consumption, making them suitable for a range of applications such as asset tracking both indoors and outdoors~\cite{zhao2021uloc, arun2022realtime}, robotics localization~\cite{ochoa2022landing,lee2022drone, zheng2022uwbgraphopt, cao2020accurate} and collaboration~\cite{hepp2016omni,queralta2022viouwbbased, corrales2008hybridtracking}, and augmented reality~\cite{molina2018augmented}.

A major challenge for accurate UWB ranging is the distortion due to obstacles in the line of sight between ranging devices.
Prior methods have addressed these by analyzing the raw channel impulse response on UWB radios~\cite{barral2019nlos,angarano2021edge, tran2022deepcir}, which is resource intensive and makes real-time operation on embedded platforms challenging.
A more common approach is sensor fusion (e.g., particle or Extended Kalman Filters, EKF) to incorporate data from other sensors such as IMUs~\cite{feng2020kalman,ochoa2022landing, mueller2015fusing, hol2009tightlycoupled}, and cameras~\cite{queralta2022viouwbbased, Xu2022omniswarm}.
We also build on an EKF-based UWB+IMU filter. 

Previous work has focused on UWB positioning via trilateration with \emph{fixed anchors} in the environment or using at least four rigidly coupled antennas on a node.
In contrast, our method solely measures inter-sensor distances, thus requiring only a \emph{single} UWB antenna and eliminating the need for stationary instrumentation. 

\section{Method}
\label{sec:method}

\subsection{Problem Statement}

Our goal is to estimate full-body pose from the sequential (raw) IMU observations and inter-sensor UWB ranging cues from a sparse set of $S$ wearable sensors.
In line with prior work, we examine this problem with 6 body-worn sensors on the pelvis, knees, wrists, and head.
Given $S$ sensor nodes with a 6-DoF IMU and a single UWB radio, we first aim to estimate each node's 3D orientation $R_t\in \mathbf{R}^{S \times 3}$, global accelerations $A_t\in \mathbf{R}^{S \times 3}$, as well as inter-sensor distances between all nodes $D_t \in \mathbf{R}^{S \times S}$. 
We then aim to predict the joint angle $\mathbf{\Theta}^i_t$ as well as the global translation $T_t$ in SMPL human model parameters~\cite{loper2015smpl}.
The pelvis sensor is the root of our body model, whereas the other sensors are leaf nodes.

\autoref{fig:pipeline} shows an overview of our pipeline as detailed below.

\subsection{Wearable Sensing Pipeline}

\subsubsection{Sensing Hardware}
Our method is designed to estimate poses from commodity sensors, i.e., those commonly found in commercial devices.
As such, we developed 6 wireless prototypes with a 6DoF IMU (LSM6DS) and a UWB radio (DWM1000) each, integrated into a 35$\times$35\,mm package (\autoref{fig:teaser} and \autoref{fig:embedded_platform}).
An onboard microcontroller (NRF52840) runs custom firmware to sample the IMU and implement a UWB ranging protocol, streaming the results over BLE to a host computer.
The host handles synchronization across the 6 devices and runs the rest of the sensing pipeline. 

\subsubsection{Obtaining Orientation and Acceleration Measurements}
We sample each  IMU at 100\,Hz to obtain sensor-frame acceleration $\mathbf{a}_i$ and angular velocity $\mathbf{\omega}_i$ signals from sensor $i$  in $S$. We implement a VQF filter~\cite{laidig2023vqf} to estimate gravity-compensated acceleration $\mathbf{\hat{a}}_i$ in the world frame and the absolute orientation quaternion $\mathbf{\hat{q}}_i$.
At every time step, the estimates from each sensor are concatenated to form the orientation and global acceleration sequences  $R_t$ and  $A_t$ respectively. 
We perform a simple IMU calibration by estimating the gyroscope and accelerometer offsets as the wearer holds an initial T-pose. 

\subsubsection{UWB Ranging}

The 6 devices implement a broadcast two-way ranging protocol to obtain the UWB distances $d_{i,j}$ between each sensor pair $(i, j)$. In a ranging round, illustrated in~\autoref{fig:ranging_protocol}, Responder devices reply in turn to a request from an Initiator with the timestamps required to resolve time-of-flight $\tau_{ij}$ between devices. 
\begin{align}
\tau_{ij}  = \frac{1}{2}(t_{i_{Received}} - t_{i_{Sent}} + t_{j_{Received}} - t_{j_{Sent}})
\label{eq:uwb_dist}
\end{align}

Initiator and Responder roles are assigned at startup. In a network of 6 devices, this protocol allows us to measure all 15 pairwise distances at a rate of 25\,Hz.

Individual UWB radios require an initial calibration to correct for hardware-specific variables, notably antenna delay which significantly affects ranging accuracy~\cite{dw_usermanual}. 
We calibrate all nodes at once, using a RANSAC regression to find an affine mapping of the raw ranges to the ground truth captured with a marker-based motion capture system. 
This calibration process is conducted once and applied consistently regardless of sensor placement or subject. The calibration mapping is however sensitive to temperature. with a variation of 2.15\,mm/°C~\cite{apnote_14}.

The parameters determined through this process remain valid given similar operating temperatures ~\cite{apnote_11}.

\subsubsection{Estimating Filtered Inter-Sensor Distance Measurements}
The raw UWB ranges exhibit increased noise when the body obstructs the direct line of sight between sensor pairs. To mitigate this, we fuse UWB ranges with the acceleration and orientation estimates for each sensor pair with an Extended Kalman Filter. 

Our filter tracks the state space comprised of relative position, relative velocity, and relative orientation between a sensor pair $(i,j)$, which we express as 
$\mathbf{x}=\begin{bmatrix}\boldsymbol{x}_{ij} &\boldsymbol{v}_{ij} & \boldsymbol{q}_{ij}\end{bmatrix}^\top$.  We initialize the state assuming a starting position in T-pose. 
In the prediction step, our filter relies on dead-reckoning using the estimated acceleration and orientation as control inputs $\mathbf{u}=\begin{bmatrix}\boldsymbol{\hat{a}}_{i} & \boldsymbol{\hat{a}}_{j} & \boldsymbol{\hat{q}}_{i} & \boldsymbol{\hat{q}}_{j}\end{bmatrix}^\top$. 
The predicted state $\mathbf{\hat{x}}_t$ at discrete time step $t$ becomes
\begin{align}
\mathbf{\hat{x}}_t=f(\mathbf{x}_{t-1}, \mathbf u_t) =
\begin{bmatrix}
\boldsymbol{x}_{ij_{t-1}} + \Delta T\boldsymbol{v}_{ij_{t-1}} + \frac{\Delta T^{2}}{2} (\boldsymbol{\hat{a}}_{j_t} - \boldsymbol{\hat{a}}_{i_t}) \\
\boldsymbol{v}_{ij_{t-1}} + \Delta T ( \boldsymbol{\hat{a}}_{j_t} - \boldsymbol{\hat{a}}_{i_t})  \\
\boldsymbol{\hat{q}_{i_{t}}}^{-1}\boldsymbol{\hat{q}_{j_{t}}}\\
\end{bmatrix} 
\label{eq:prediction_step}
\end{align}
where $\Delta T$ is the difference between consecutive time steps.
 
The noise in the prediction step is mainly introduced by the control input and therefore not simply additive. 
The VQF filter lacks explicit orientation covariance outputs. 
Instead, we treat we $\boldsymbol{\hat{a}}$ and $\boldsymbol{\hat{q}}$ inputs as independent, and define the prediction step noise covariance $\boldsymbol{Q}_t$ as:
\begin{align}
\boldsymbol{Q}_t = \boldsymbol{W}_t\Sigma_u\boldsymbol{W}_t^T
&  &
\boldsymbol{W}_t = \frac{\partial f(\mathbf{x}_{t-1}, \mathbf u_t)}{\partial\mathbf u}
\end{align}
where $\boldsymbol{\Sigma}_u$ is the diagonal matrix of each of the input sensors' variance, which we determine from ground-truth.
We introduce UWB ranges at the correction step.  
Before fusion, we remove raw UWB outliers outside a range of acceptable inter-sensor distances, to ignore measurements that are incompatible with human motion constraints. 
This range for each sensor pair is defined apriori assuming body heights of $[1.5, 2\,\text{m}]$ and constant across subjects.
These ranges and their first derivative, form the measurement model $\mathbf{h}(\mathbf{x})$ defined in Eq.\,\ref{eq:h}, and linearized as Eq.\,\ref{eq:H}.
\begin{align}
\mathbf{h}_t(\mathbf{x}) & =   
\begin{bmatrix}
d \\
v \\
\end{bmatrix} = 
\begin{bmatrix}
\| \boldsymbol{x}_{ij} \|_2 \\
\| \boldsymbol{v}_{ij} \|_2 \\
\end{bmatrix}
\label{eq:h}
\\
H(\mathbf{x}) & = \frac{\partial h(\mathbf{x})}{\partial\mathbf x} =
\begin{bmatrix}
\frac{\boldsymbol{x}_{ij}}{x}&\boldsymbol {0}_3 \\
\boldsymbol{0}_3 & \frac{\boldsymbol{v}_{ij}}{v} \\
\end{bmatrix}_{2\times6}
\label{eq:H}
\end{align}
The measurement covariance matrix is simply the diagonal matrix of the distance and speed variances observed from raw distance measurements against ground-truth. It is different for each sensor pair, as some are LOS while others are not, but constant across subjects.
Without direct observations, the relative position estimate  $\boldsymbol{x}_{ij}$ is still prone to drift, however, its norm $\boldsymbol{\hat{d}}_{ij} = \| \boldsymbol{x}_{ij} \|_2$ provides a distance estimate more robust to NLOS distortions.  

\begin{figure}[b]
    \centering
    \includegraphics[width=\linewidth]{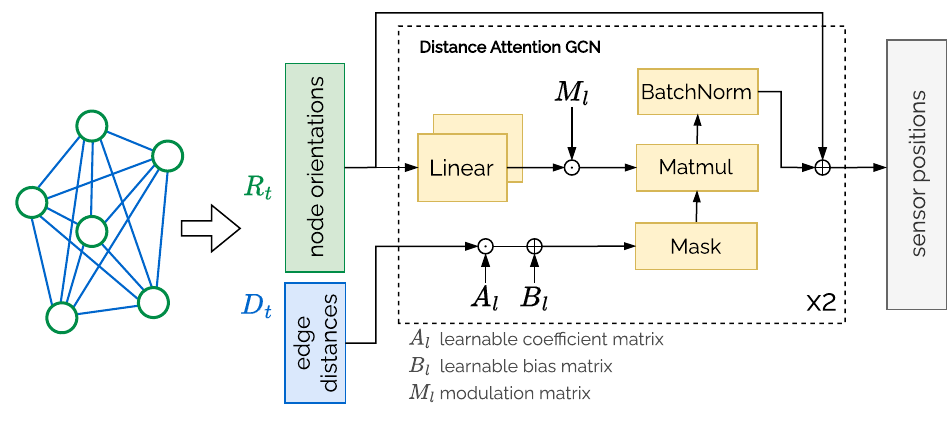}
    \caption{Architecture details for the Distance Attention Graph Convolutional Network (DA-GCN). Given node orientation $R_t$ and edge distance $D_t$, DA-GCN estimates sensor positions conforming to the distance measurements. The model consists of a distance attention branch with learnable matrices $A_l$ and $B_l$ to represent the correlation between nodes and a weight-modulation branch $M_l$ to learn disentangled transformation for different nodes.}
    \label{fig:network_detail}
    \Description{The figure illustrates our Distance Attention Graph Convolutional Network module. On the left, a graph with six nodes, outlined in green and holding orientation information $R_t$, with blue edges between all nodes indicating distance information $D_t$. This graph serves as input to our GCN. The edge distances are multiplied by a learnable coefficient matrix $A_l$ and added to a learnable bias matrix $B_t$, then passed through a Mask operation. The node orientations go through two linear layers and are multiplied by a Modulation matrix $M_l$. Afterward, this result is matrix-multiplied by the masked distances and batch normalized. The orientation data is added back at the end of the process. The sequence is performed twice to estimate sensor positions.}
\end{figure}

\subsection{Pose Estimation with Distance Constraints}

\subsubsection{Model}

The first module of our estimation pipeline is a sensor position estimator. It consists of two branches to capture the temporal and spatial information from input signals. In the first, we deploy an LSTM network~\cite{hochreiter1997long} cells to estimate sensor positions relative to the root from acceleration and orientation estimates $A_t$ and $R_t$. This module leverages temporal information from the sequence of inputs, and we denote its output as $\mathbf{P}_{t} \in \mathbf{R}^{S \times 3}$. 

In parallel, we introduce a second branch to estimate relative sensor positions from pairwise distances using a Distance Attention Graph Convolutional Network (DA-GCN). 
As illustrated in ~\autoref{fig:network_detail}, we represent our system of wearable devices with a graph $\mathbf{G} = \{\mathbf{V},\mathbf{E}\}$, with vertices $\mathbf{V} \in S$  and edges $\mathbf{E}$. 
Each node $i$ holds the orientation estimate $\mathbf{R}_{t}^{i}$, while edges represent inter-sensor distance estimates $\mathbf{D}_{t}$. 
Previous GCN implementations~\cite{zhao2019semantic} exploit a pre-defined affinity matrix with binary values to represent connectivity between joints of a skeleton model. 
We take it further by encoding inter-sensor distances on edge features, to capture additional correlation between nodes, to which we apply a modulated graph convolutional network~\cite{zou2021modulated}. 
More specifically, we first normalize inter-sensor distance by the distance between the head and pelvis to rule out differences in human shape. 
Building on the success of channel attention in image processing~\cite{woo2018cbam,jiang2021towards}, we then compute the correlation matrix as in~\autoref{eq:correlation_matrix}, where elements in matrix $A_l$ and $B_l$ are learnable.
\begin{align}
\boldsymbol{C}_{l}=\boldsymbol{A_l} \odot \boldsymbol{D} \oplus \boldsymbol{B_l}
\label{eq:correlation_matrix}
\end{align}
The correlation matrix is then applied as a weight to aggregate modulated node features, which can be denoted as:
\begin{equation}
\boldsymbol{H}_{l+1}=(\boldsymbol{M}_{l} \odot (\boldsymbol{W}_{l}\boldsymbol{H}_{l}))  \boldsymbol{C_{t}}
\label{eq:hidden_feature}
\end{equation}
, where $H_{l}$ represents the node feature at layer $l$ and $M$ represents the modulation for weight $W_{l}$ to learn disentangled transformation for node feature. By leveraging orientation and distance constraints, our DA-GCN outputs spatial sensor position estimation 
$\mathbf{P}_{s}  \in \mathbf{R}^{S \times 3}$.

Similarly to TransPose~\cite{yi2021transpose}, we use linear interpolation to fuse the two relative sensor position estimates.  
We compute a weighted sum of $\mathbf{P}_{t}$ and $\mathbf{P}_{s}$ based on the magnitude of acceleration estimates in $A_t$. 
This is driven by the fact that in scenarios with little movement IMUs have a low signal-to-noise ratio, and distinguishing between poses solely from inertial data becomes challenging.

Therefore, the fusion algorithm can be formulated as:

\begin{equation}
    \mathbf{P}^{i} = {
    \begin{cases}
    \mathbf{P}_{s}^{i}  & \| \mathbf{\hat{a}}_{i}\| \leq \underline{\mathbf{a}}\\
   \frac{ \| \mathbf{\hat{a}}_{i}\| - \underline{\mathbf{a}}}{\overline{\mathbf{a}} -  \underline{\mathbf{a}}}  \mathbf{P}_{t}^{i} +
   \frac{  \overline{\mathbf{a}} - \| \mathbf{\hat{a}}_{i}\|}{\overline{\mathbf{a}} -  \underline{\mathbf{a}}} \mathbf{P}_{s}^{i}
   
   &\underline{\mathbf{a}} < \| \mathbf{\hat{a}}_{i}\| < \overline{\mathbf{a}}\\
    \mathbf{P}_{t}^{i}  & \| \mathbf{\hat{a}}_{i}\| \geq  \overline{\mathbf{a}}
    \end{cases}
    }
\end{equation}
We set the thresholds $\overline{\mathbf{a}}$ and $\underline{\mathbf{a}}$ to 8.0 and 2.0\,m/s$^2$, respectively, which we determined via grid search to minimize position error. 

This fusion approach accommodates different update rates between the inertial (100\,Hz) and distance (25\,Hz) measurements, eliminating the need for resampling or interpolation. 

The rest of the pipeline processes the combined position estimates, acceleration, orientation, and distance measurements. These inputs are passed to a kinematics decoder, which estimates local orientation, local velocity, and foot contact, followed by a dynamics optimizer, based on previous work~\cite{yi2022physical}. The pipeline outputs SMPL pose parameters $\Theta_t$ global translation $T_{t}$.

\subsubsection{Loss Functions}

To supervise our sensor position estimator, we design a loss function that incorporates distance constraints. 
From the estimated sensor positions $\mathbf{P}_{s}$ or $\mathbf{P}_{t}$, which we denote as $\mathbf{\widehat{P}}$, we compute inter-sensor distances $d(\mathbf{\widehat{P}}) \in \mathbf{R}^{S \times S}$  and the L1 reconstruction loss $\| d(\mathbf{\widehat{P}}) - \mathbf{D} \|_{1}$. We also pose additional constraints on its direction by computing the cosine similarity between estimated sensor positions and nearest joint positions. We can then formulate the full distance-aware loss as:
\begin{equation}
    \mathbf{L}_{d} = \frac{\mathbf{\widehat{P}}\cdot\mathbf{\widetilde{P}}}{\|\mathbf{\widehat{P}}\| \|\mathbf{\widetilde{P}}\|} + \lambda\| d(\mathbf{\widehat{P}}) - \mathbf{D} \|_{1}
\end{equation}
where $\mathbf{\widetilde{P}}$ is the approximate sensor position from nearest joints. We select $\lambda$ as 0.01 and leverage this distance aware loss as the objective function for sensor position estimation. 

For other modules, we apply L2 loss for the rotation estimator, binary cross entropy loss for the contact estimator, and accumulated loss~\cite{yi2021transpose} for the velocity estimator. 

\subsubsection{Data Synthesis}
Because training our pose estimation pipeline requires a large-scale dataset with IMU+UWB measurements and SMPL references, we first appropriate AMASS as an existing motion capture dataset~\cite{AMASS:ICCV:2019} to synthesize sensor data.

Similarly to previous research~\cite{yi2021transpose}, we synthesize global acceleration and orientation data by placing virtual sensors on SMPL mesh vertices. We deduce the rotation matrix from SMPL joint rotations and derive world-frame acceleration vectors using finite differences in vertex positions, 
\begin{equation}
    \mathbf{a}_t^i = \frac{n\mathbf{p}_{t+n}^i - 2\mathbf{p}_{t}^i +  \mathbf{p}_{t-n}^i}{(n\mathbf{\Delta}t)^2}
\end{equation}
where ${p}_{t}^i$ is the position of vertices $i$ at time step $t$. 
To synthesize UWB data, we first compute the Euclidian distance from the virtual sensors' positions and then apply a time-varying noise model that reflects observations on real data. 
We conducted NLOS noise modeling experiments, in which we varied distances and obstacle sizes between sensors.
We observed a noise standard deviation similar to LOS for obstacle-to-distance ratio smaller than 0.2, and larger for ratios above 0.2.
As such, we model noise as dynamic Gaussian noise with standard deviation $\mathbf{\sigma}(t)_{ij}$ given by:
\begin{align}
\mathbf{\sigma}(t)_{ij} = {
    \begin{cases}
    \mathbf{\sigma}_0  & c(t)_{ij} < 0.2\\
    \mathbf{\sigma}_1  & c(t)_{ij} \geq 0.2
    \end{cases}
    }  
    \vspace{-2ex}
\end{align}
where $c(t)_{ij}$ is the proportion of occlusion on the line between devices $i$ and $j$. 
To compute this occlusion, we leverage the implicit human body model COAP~\cite{mihajlovic2022coap} and query sampled points along the line segment at a constant resolution between devices.
The values of $\mathbf{\sigma}_0$ and $\mathbf{\sigma}_1$ are determined empirically, by running our sensing pipeline with two devices first in LOS  ($\mathbf{\sigma}_0 = 0.051$~~m) then in NLOS with a person as obstacle ($\mathbf{\sigma}_1 = 0.083$~m).

\subsubsection{Training Details}
We adopt the Adam solver~\cite{kingma2015adam} with batch size 256  to optimize the parameters of model. We train each module separately with the learning rate at $1 \times 10^{-3}$ and decays by a factor of 0.33 every 20 epochs. We train our model with PyTorch on one NVIDIA GeForce GTX 3090 GPU. It takes about 4 hours in total to train our model.

\section{Dataset}
\label{sec:dataset}

\begin{table*}
\centering
\small 
\setlength{\tabcolsep}{2pt}
\renewcommand{\arraystretch}{1.2}
\caption{Comparison with state-of-the-art methods on existing datasets augmented with simulated, ideal inter-sensor distances}
\vspace{-3ex}
\begin{tabular}{@{}l ccc ccc ccc ccc ccc ccc@{}} 
    \multicolumn{1}{c}{} & \multicolumn{9}{c}{original TIP/PIP training set (TotalCapture ground truth included)} & \multicolumn{9}{c}{AMASS \emph{without} TotalCapture} \\
    \cmidrule(l{2pt}r{3pt}){2-10} \cmidrule(l{3pt}r{2pt}){11-19}
    Dataset & \multicolumn{3}{c}{DanceDB} & \multicolumn{3}{c}{DIP-IMU} & \multicolumn{3}{c}{TotalCapture} & \multicolumn{3}{c}{DanceDB} & \multicolumn{3}{c}{DIP-IMU} & \multicolumn{3}{c}{TotalCapture} \\
    \cmidrule(l{2pt}r{2pt}){2-4} \cmidrule(l{2pt}r{2pt}){5-7} \cmidrule(l{2pt}r{3pt}){8-10} \cmidrule(l{3pt}r{2pt}){11-13} \cmidrule(l{2pt}r{2pt}){14-16} \cmidrule(l{2pt}r{2pt}){17-19}
    Method & SIP Err & Pos Err & Jitter & SIP Err & Pos Err & Jitter & SIP Err & Pos Err & Jitter & SIP Err & Pos Err & Jitter & SIP Err & Pos Err & Jitter & SIP Err & Pos Err & Jitter \\
    \midrule
    TIP & 18.705 & 8.500 & 1.438 & 17.068 & 5.822 & 0.882 & 11.361 & 5.145 & 0.751 & 18.740 & 8.501 & 2.252 & 17.011 & 5.710 & 1.162 & 13.588 & 5.967 & 1.015 \\
    PIP & 20.007 & 8.877 & 2.234 & 15.020 & 6.020 & 0.240 & 12.930 & 7.025 & 0.204 & 20.007 & 8.997 & 2.334 & 15.977 & 6.209 & 0.260 & 15.930 & 7.046 & 0.292 \\
    \multicolumn{4}{l}{with synthesized \emph{perfect} distances:} &&& &&& &&& &&& \rule{0pt}{2.5ex}\\
    TIP-D & 16.243 & 7.624 & 2.220 & 15.910 & 5.442 & 0.863 & 10.762 & 5.142 & 0.722 & 16.225 & 7.648 & 2.221 & 15.914 & 5.259 & 1.128 & 12.183 & 5.507 & 1.004 \\
    PIP-D & 16.205 & 8.042 & 2.124 & 13.762 & 5.336 & 0.281 & 11.408 & 5.560 & 0.201 & 16.815 & 8.063 & 2.166 & 13.788 & 5.355 & 0.297 & 13.161 & 6.317 & 0.273 \\
    \textbf{UIP (ours)} & \textbf{15.280} & \textbf{7.450} & \textbf{0.430} & \textbf{13.200} & \textbf{5.050} & \textbf{0.240} & \textbf{10.707} & \textbf{5.108} & \textbf{0.206} & \textbf{15.321} & \textbf{7.612} & \textbf{0.430} & \textbf{13.210} & \textbf{5.053} & \textbf{0.248} & \textbf{11.321} & \textbf{5.490} & \textbf{0.257}
\end{tabular}
\vspace{1ex}
\label{table:existing_data}
\vspace{-2ex}
\end{table*}

\begin{table*}[ht]
\vspace{1ex}
\setlength{\tabcolsep}{1.0pt}
\small
\raggedright
\parbox[t][][t]{.50\linewidth}{
\caption{Comparisons on UIP-DB with acceleration, orientation and inter-sensor distances from real off-the-shelf sensors.}
\vspace{-3ex}
\begin{tabular}{@{}l ccc ccc ccc@{}}
     UIP-DB & \multicolumn{3}{c}{overall}& \multicolumn{3}{c}{$\|\mathbf{a}\| \leq 1.0 \, \text{m/s}^2$} & \multicolumn{3}{c}{$ \|\mathbf{a}\| > 1.0 \, \text{m/s}^2$} \\
   \cmidrule(l{2pt}r{2pt}){2-4} \cmidrule(l{2pt}r{2pt}){5-7} \cmidrule(l{2pt}r{2pt}){8-10}
    Method & SIP Err & Pos Err & Jitter & SIP Err & Pos Err & Jitter & SIP Err & Pos Err & Jitter \\
    \midrule
    TIP & 33.01  & 14.82 & 1.86 & 36.90 & 16.55 & 1.79 & 28.83 & 12.95 & 1.95 \\
    PIP & 30.47  & 13.62 & 1.57 & 36.18 & 15.87 & 1.33 & 23.18 & 10.80 & 1.82 \\
    \multicolumn{7}{l}{with distances \emph{estimated} from UWB recordings:} \rule{0pt}{2.5ex}\\
    TIP-D & 30.34  & 13.96 & 1.84 & 33.11 & 15.52 & 1.79 & 27.17 & 12.19 & 1.88 \\
    PIP-D & 30.33  & 13.27 & 1.39 & 35.47 & 15.25 & 1.29 & 24.13 & 10.80 & 1.51 \\
    \textbf{UIP (ours)} & 24.12  &  10.65 & \textbf{0.05}& 24.72 & 11.84 & \textbf{0.051} & 22.64 & 10.02 & \textbf{0.06} \\
     \textbf{UIP finetuned} & \textbf{23.85} &    \textbf{10.55}& 0.08&  \textbf{24.51}& \textbf{11.18} & 0.078 & \textbf{22.43} &\textbf{ 9.79} & 0.08\\
\end{tabular}
\label{table:our_data}
}
\hspace{0.03\linewidth}
\parbox[t][][t]{.45\linewidth}{
\centering
\setlength{\tabcolsep}{8.0pt}
    \caption{Results of our ablation studies. \\ ~}
    \vspace{-4ex}
    \begin{tabular}{lccc}
       \rule{0pt}{3ex} 
    Method & SIP Err & Pos Err & Jitter\\
    \midrule
    \textbf{UIP (ours)}& \textbf{ 24.115} & \textbf{ 10.648} & \textbf{0.055}\\
    w/o inter-sensor distances & 31.008 & 13.749  &  1.556 \\
    w/o IMU  & 27.005   & 14.086 & 0.098  \\
    w/o DA-GCN & 27.529  &  12.136 & 0.092 \\
    w/o UWB noise & 27.006  & 12.440 & 2.125  \\   
    w/~~~\,\,\,Gaussian UWB noise & 26.068  &  12.159 & 2.005\\ 
    w/o Distance-guided loss & 25.830  & 11.777 & 1.966\\
    \end{tabular}
    \vspace{0.5ex}
    \label{table:ablations}
}
\vspace{-3ex}
\end{table*}

To evaluate our method and to spur further work, we collected a motion capture dataset of a large variety of movement types, capturing the raw IMU+UWB measurements of our tracker nodes alongside those from an optical tracking system.
We recorded more than 25 different movements from 10 participants (6 male, 4 female), with heights between 155\,cm and 187\,cm.

Our recording study comprised two sessions.
The first focused on everyday movements that participants performed standing up, such as walking, jumping, squats, etc.
The second session involves interactions with a chair and movements in various seated positions.
For each participant, we recorded two takes of each session and a final freestyle session, where they acted as desired for 4--6\,min.

\begin{figure*}
\setlength{\tabcolsep}{1.0pt}
\small
\raggedright
\parbox[t][][t]{.48\linewidth}{
    \centering
    \includegraphics[width=\linewidth]{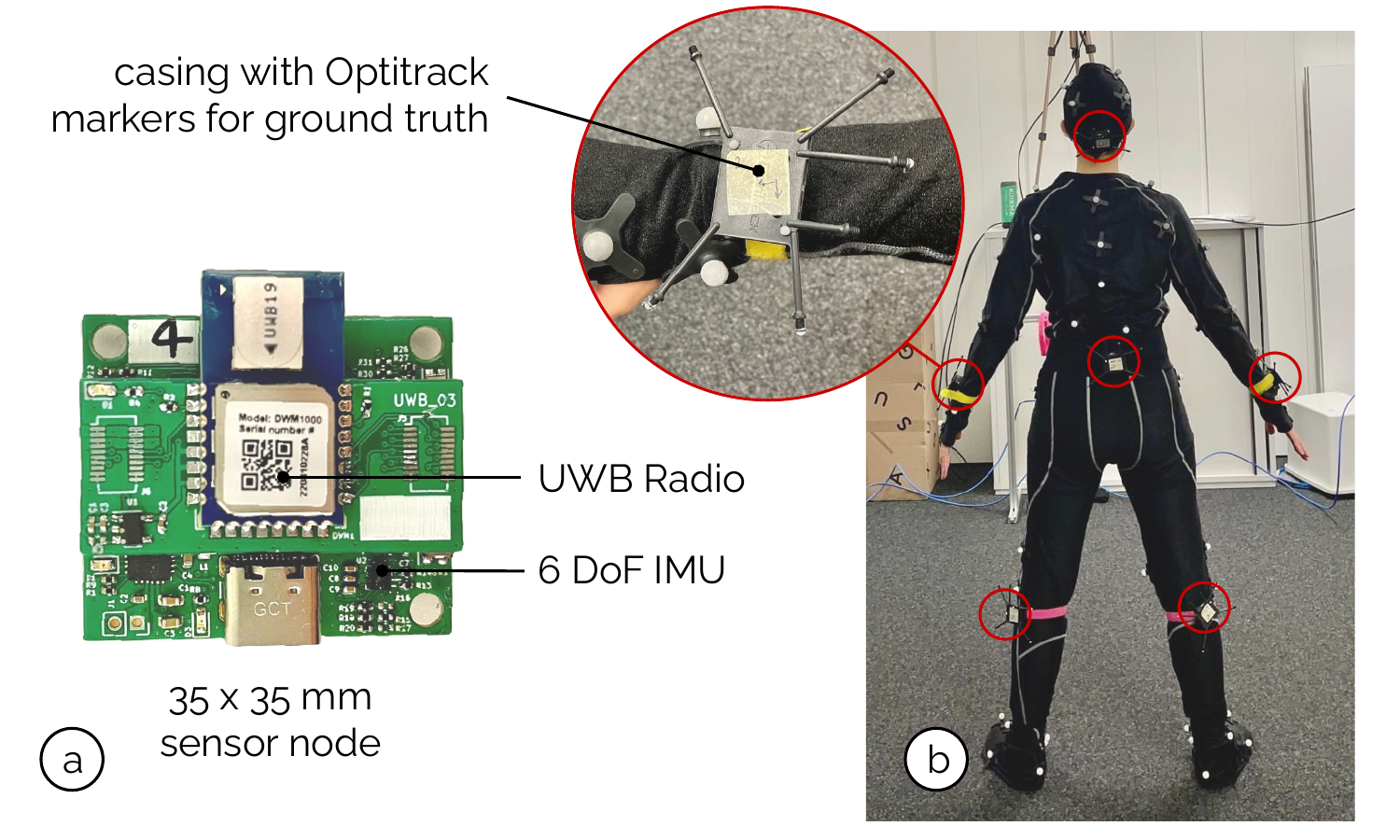}
    \caption{(a)~Our embedded sensor nodes and
    (b)~a person wearing six of them inside a 20-camera Optitrack motion capture setup for reference poses.}
    \Description{The figure comprises two parts labeled (a) on the left and (b) on the right, presenting our sensor system for motion tracking. Part (a) shows a close-up of a sensor node that measures 35 by 35 millimeters. This node includes a 6 degrees of freedom IMU and an Ultra-Wideband Radio module, which are mounted on a circuit board. Part (b) features an individual wearing a black motion capture suit with reflective markers used to capture reference SMPL poses. The person is also wearing 6 sensor nodes, on the wrists, knees, head, and lower back highlighted with a red circle. The inset in part (b) is a zoomed-in view of one of the sensors in a case with reflective markers used to track its ground-truth position and orientation.}
    \label{fig:embedded_platform}
}
\hspace{0.03\linewidth}
\parbox[t][][t]{.48\linewidth}{
 \centering
    \includegraphics[width=\linewidth]{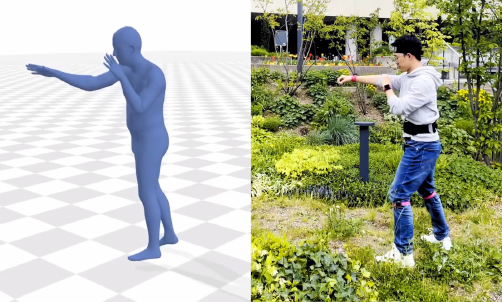}
    \caption{Example of motion tracking using our method of a person exercising outdoors, where no tracking infrastructure is needed for reference or anchoring.}
    \label{fig:video}
    \Description{This figure shows a person exercising outdoors and the corresponding pose estimated by our UIP method side-by-side.}
}
\end{figure*}

As shown in \autoref{fig:embedded_platform}, participants wore a suit with 57 reflective markers and 6 of our wearable devices on the back of the head, lower back, wrists, and knees.
The wearable sensors continuously collected raw 3D accelerometer, 3D gyroscope, and 3D magnetometer readings at 100\,Hz as well as raw UWB ranges at 25\,Hz.
We extended the recordings with the filtered estimates of global acceleration, orientation, and inter-sensor distances, estimated using our processing pipeline as described above.

We obtained ground truth for the poses of wearable sensors and body joints from a 20-camera Optitrack system across a capture area of $4\times5$\,m.
To extract SMPL references and global translation, we labeled the point cloud from the marker suit with SOMA~\cite{SOMA:ICCV:2021}, and then used MoSh++~\cite{AMASS:ICCV:2019}.
Since each of our trackers rigidly mounted a reflective marker, our dataset also includes their reference positions and orientations. 

Each recording session started and ended with a T-pose to calibrate IMU readings and initialize the inter-sensor distances EKF.
A jump then synchronized all sensors with the extracted SMPL poses.

\section{Experiments}
\label{sec:experiments}

We analyze our pipeline's accuracy, visually and quantitatively compare it to existing baselines, and discuss ablation studies.

\subsection{Sensing Pipeline Evaluation}
\label{sec:evaluation_sensors}

From the ground-truth sensor orientations and positions recorded in our dataset, we derive reference inter-sensor distances and accelerations. We use these values to evaluate our sensing pipeline and quantify the error it introduces at the input of pose estimation.

Our orientation estimate drifts at a rate of 3.21\textdegree/min, 0.42\textdegree/min and 0.31\textdegree/min for roll, pitch, and yaw angles respectively, in line with results obtained from similar commodity IMUs in previous work~\cite{laidig2023vqf}.
As this estimate is also used to transform sensor-frame acceleration into the global frame and filter UWB ranges, our dataset provides less accurate model inputs than those in existing datasets~\cite{trumble2017total,huang2018deep} that rely on proprietary drift-free estimators in high-end sensors. 

\begin{figure}[t]
    \centering
    \includegraphics[width=\linewidth]{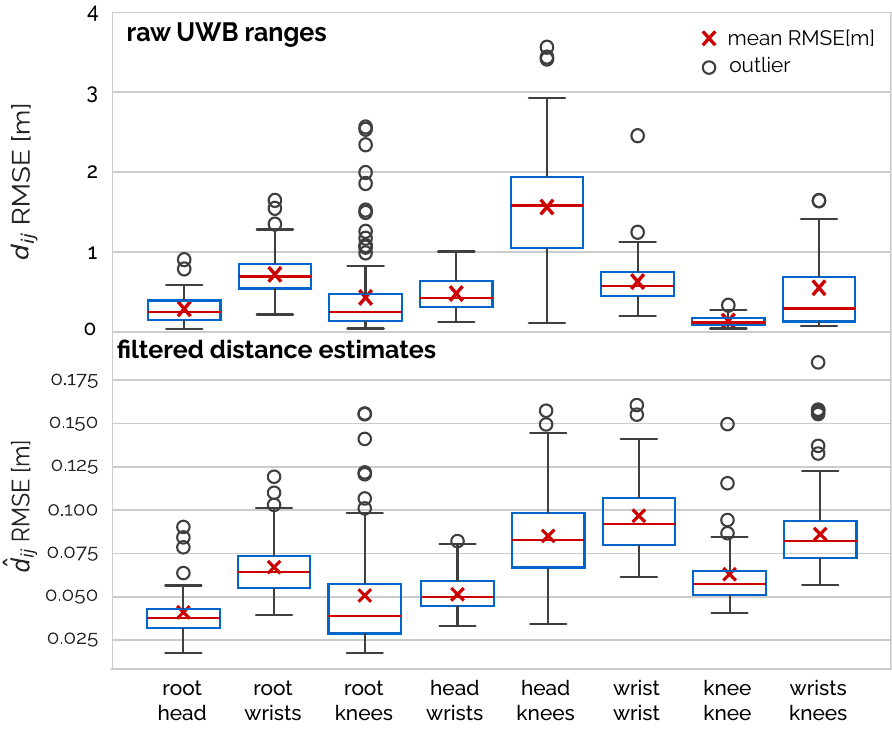}
    \caption{Inter-sensor distance RMSE in our dataset.}
    \Description{The image contains two vertically stacked boxplots of the Root Mean Square Error in our UIP-DB dataset. The top plot shows the RMSE of raw UWB ranges between 0 and 4 meters, and the bottom one shows the RMSE of inter-sesnor distance estimates after filtering, ranging from 0 to 0.175 meters. Both plots contain side-by-side box-and-whisker diagrams for different sensor pairings: root-head, root-wrists, and root-knees. Each box plot shows the median, quartiles, and outliers, which are marked as individual circles. Red 'X' marks indicate the mean RMSE for each pairing. Both plots reveal a general trend: the RMSE is higher for sensor pairs with larger body occlusion, and the overall error is greatly reduced by our filtering pipeline.}
    \vspace{-1ex}
    \label{fig:uwb_rmse}
\end{figure}

\autoref{fig:uwb_rmse} shows the distribution of the RMSE of the pairwise distances estimated by our filtering pipeline.
The average error is largest for wrist--wrist and head--knees sensor pairs, as the body frequently occludes the line of sight between sensors.

\subsection{Pose Estimation Evaluation}

\subsubsection{Evaluation protocol}
We compare our approach against state-of-the-art IMU-based pose estimation methods: PIP~\cite{yi2022physical} and TIP~\cite{jiang2022transformer}.
For fair comparison, we also augment both TIP and PIP by concatenating pairwise distances to their original inputs and retrain them using publicly available resources.

Our first experiment shows the benefits of adding pairwise distances, evaluating all methods on existing datasets.
We form a test set from DIP-IMU~\cite{huang2018deep}, TotalCapture~\cite{trumble2017total}, and DanceDB---held out from the AMASS training set.
The acceleration and orientation measurements in TotalCapture and DIP-IMU were captured from XSens suits, while DanceDB has synthetic values.
We add ideal synthetic inter-sensor distances to our test set to study their contribution independent of noise. 

We first train all methods on AMASS, as originally done in TIP and PIP, with synthetic acceleration, orientation, and inter-sensor distances with our time-varying noise model.
However, because this training set includes synthesized sensor readings from TotalCapture's ground truth, we additionally run these experiments with models trained on AMASS \emph{without} TotalCapture.

A second experiment evaluates performance on our collected dataset UIP-DB, examining results both overall and for a mean acceleration split: slow movements (mean acceleration below 1 m/s²) and fast movements (mean acceleration above 1 m/s²).
All inputs are from actual measurements processed by our pipeline.

\subsubsection{Metrics}
We evaluate model performance on three metrics.
SIP Error measures the mean orientation error of the upper arms and legs in the global space in degrees.
Joint position error measures the mean joint distance between the reconstructed and ground-truth joints with both root position and orientation aligned in cm.
Jitter is the mean jerk (time derivative of acceleration) of all body joints in the global space in $km/s^3$, reflecting motion smoothness.

\subsubsection{Results}

\autoref{table:existing_data} lists the evaluation results on DanceDB, TotalCapture, and DIP-IMU datasets with perfect synthetic inter-sensor distances.
\autoref{table:our_data} has the results on our captured dataset.

Methods with distance constraints 
perform consistently better than inertial-only pipelines, with the most significant improvements in the SIP angle error.
Within distance-augmented methods, our approach shows a moderate advantage when evaluated on existing datasets.
However, its performance is markedly better on real data from our dataset, with 20\% lower SIP and position errors, and an 80\% reduction in jitter compared to distance-augmented PIP. 

\begin{figure*}
\setlength{\tabcolsep}{1.0pt}
\small
\raggedright
\parbox[t][][t]{.48\linewidth}{
    \includegraphics[width=\linewidth]{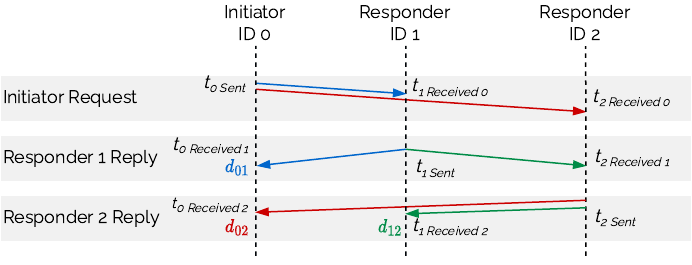}
    \caption{Ranging transaction with three devices. Timestamps to resolve time-of-flight are included in the UWB message payload and thus broadcast to all network participants.}
    \Description{
    The figure is a timing diagram that illustrates communication between an initiator and two responders, labeled as ID 0, ID 1, and ID 2, respectively. The first message in the protocol is the "Initiator Request" sent from the initiator (ID 0) to both responders. The request starts at a time $t_{0\text{Sent}}$. It is received by Responder 1 at $t_{1\text{Received_{0}}}$ and by Responder 2 at $t_{2\text{Received_{0}}}$. After a delay, Responder 1 sends its reply at time $t_{1\text{Sent}}$, which reaches the Initiator at $t_{0\text{Received_{1}}}$. At this point, the system can determine the distance $d_01$ between the Initiator and responder 1. Responder 2 receives Responder 1's reply at $t_{2\text{Received_{1}}}$, after which it sends its own reply at $t_{2\text{Sent}}$. Responder 2's reply reaches Responder 1 at $t_{1\text{Received_{2}}}$, from which distance between Responders 1 and 2 $d_12$ is calculated. It reaches the Initiator at $t_{0\text{Received_{2}}}$, at which point the distance between Initiator and Responder 2 $d_02$ can be determined.}
    \label{fig:ranging_protocol}
}
\hspace{0.03\linewidth}
\parbox[t][][t]{.48\linewidth}{
\centering
\includegraphics[width=\linewidth]{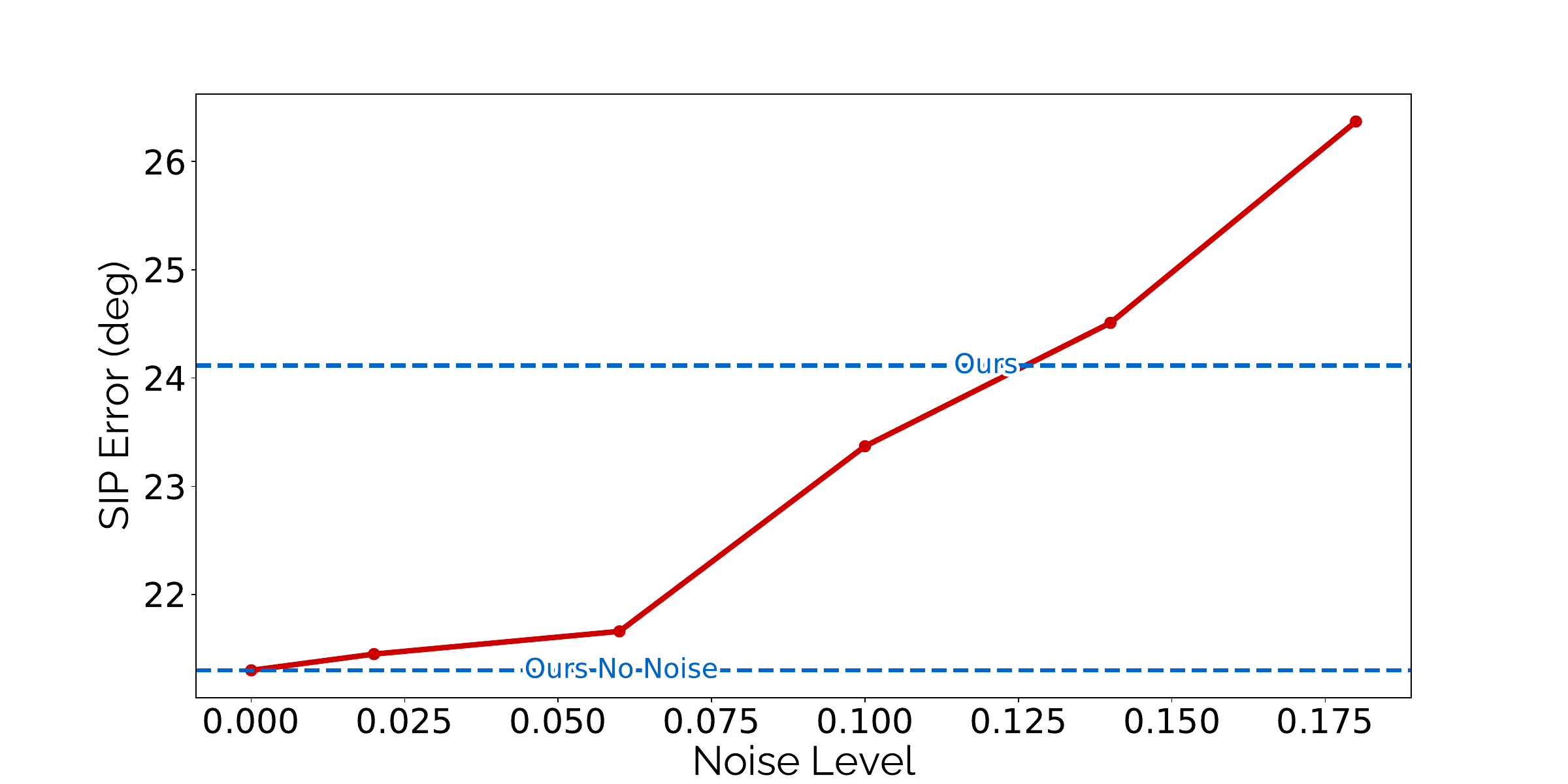}
    \caption{SIP error of UIP on our dataset at different levels of synthetic Gaussian noise on inter-sensor distances.}
    \label{fig:eval_noise_level}
    \Description{ This figure is a graph plotting the SIP Error in degrees on the vertical axis against the noise level, or its standard deviation,  on the horizontal axis. The SIP error plot in red increases as the noise level rises. Two dashed blue lines indicate the performance levels of the system without noise ("Ours-No-Noise"), for which the SIP Error reaches 21 degrees, and at the average noise level we observed in our recorded dataset or just over 24 degrees.}
}
\end{figure*}

This disparity is due to two main factors.
First, our evaluation uses \emph{inexpensive off-the-shelf} sensors, such that the acceleration, orientation, and inter-sensor distance estimates from our dataset exhibit more noise and drift than synthetic datasets or those from high-end sensors.
This also partially explains the larger error obtained by TIP and PIP on our dataset.
In contrast, our fusion pipeline and proposed loss function actively enforces distance constraints while adaptively introducing constraints from inertial data, showing better results in all presented datasets.
\autoref{fig:eval_noise_level} compares how varying levels of noise in the distance estimate influence SIP error.

\begin{figure}
    \centering
    \includegraphics[width=\linewidth]{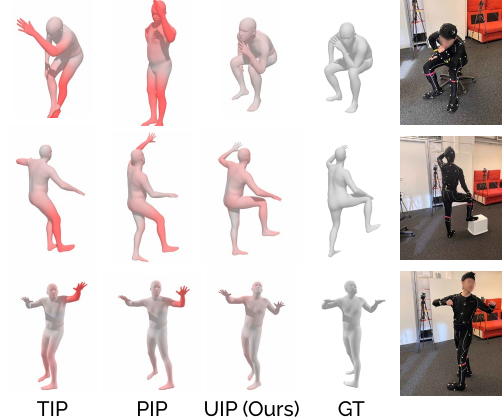}
    \caption{Qualitative comparison of pose estimates from the recordings in our dataset.
    \autoref{fig:compare_long} shows more examples.}
    \Description{ The figure presents a grid of SMPL avatars, for a visual comparison of pose estimation results for different poses and methods. The grid has 5 columns and 3 rows. The left-most column corresponds to pose estimates from TIP, the next to estimates from PIP, then our method UIP followed by ground truth. The right-most column shows photographs of the subject performing the reconstructed poses while wearing a motion capture suit. Each row illustrates a different pose. The top row shows a sitting pose where the subject leans forward with his elbow on his knee. The middle row shows a pose in which the subject rests his right leg on an elevated. In the bottom row, the subject rotates about their spine with their arms parallel to the ground. The various SMPL reconstructions are grey with red highlights on the limbs, indicating estimating errors. Overall, pose estimates from TIP and PIP exhibit more pronounced errors, especially around the arms and legs.}
    \label{fig:qualitative_results}
\end{figure}

\begin{figure*}
    \centering
    \includegraphics{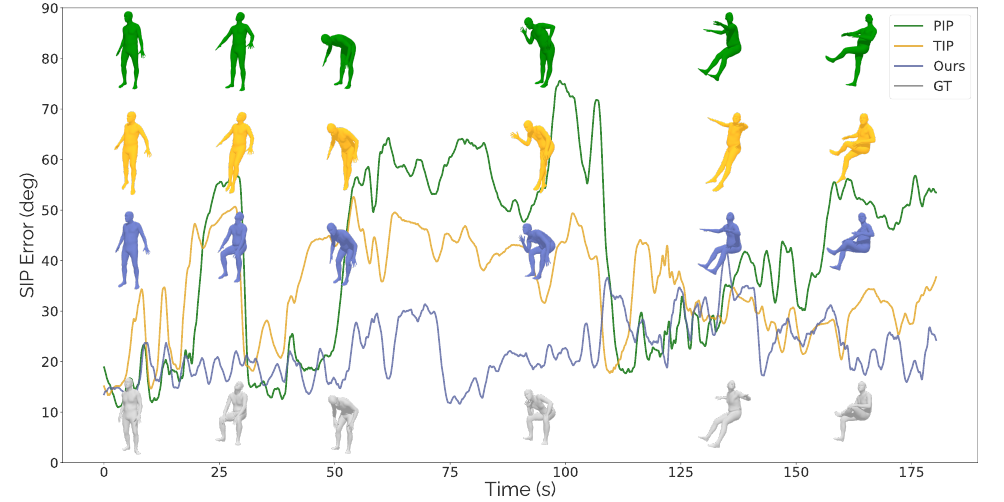}
    \vspace{-2mm}
    \caption{Visualization of SIP Error over time on a test sequence from UIP-DB.}
    \Description{
    This figure plots the SIP Error in degrees on the vertical axis over time in seconds on the horizontal axis. The plot corresponds to a motion sequence with alternating standing and sitting poses over a duration of 180 seconds. There are four lines, each representing different benchmarks, with matching pose representations for each method at specific timestamps. Green line and avatars represent PIP, TIP is in yellow, Our method UIP is in blue and ground truth is in grey. The plot shows that while the SIP error for UIP fluctuates over time, it does so much less than that of TIP and PIP, which have notable peaks in segments corresponding to seated poses.
    }
    \label{fig:error_v_time}
    \vspace{-2ex}
\end{figure*}

Second, our dataset contains many sequences that are difficult to disambiguate from inertial data, such as motions with low speed. 
\autoref{table:our_data} shows that TIP and PIP---based solely on IMUs---perform significantly worse for small accelerations, whereas performance improves for quick motions.
Our method effectively captures sitting poses and interactions with objects (\autoref{fig:qualitative_results}).
\autoref{fig:error_v_time} plots the SIP error over a long sequence while a participant alternated between standing and sitting. 
Our method consistently tracks these state changes, whereas TIP and PIP incorrectly revert to a standing pose.

\subsubsection{Ablations}
We perform ablation studies on our dataset to understand the impact of each of the components we propose in our method, with results summarized in~\autoref{table:ablations}.
We first assess the individual contribution of each of the inertial and distance inputs to the pose estimation pipeline, by excluding them in turn. We observe that distance constraints play a more important role in reducing SIP angle error and jitter, while inertial inputs have a more significant impact on the joint position error metric.

Next, we determine the effectiveness of our DA-GCN module.
Our ablation discards DA-GCN and directly concatenates inter-sensor distances to the inertial data as input to the LSTM.
This variant only considers temporal information and ignores the spatial correlation between nodes, leading to higher SIP and position errors.
Investigating the effect of our collision-aware UWB noise model for training, we first train with ideal synthesized distances, then with simple additive Gaussian noise ($\sigma = 0.12$).
SIP error and jitter metrics are impacted most by this ablation, showing the importance of a representative noise model in training our method. 

Our final ablation study discards the pairwise distance reconstruction term from our loss function and substitutes cosine similarity with an L2 loss.
This affects jitter most, indicating that the loss primarily contributes to stabilizing our predictions. 
 
\section{Limitations and Future Work}
\label{limitations}

By incorporating pairwise distances with inertial measurements, our system adds a requirement for sensing input that is UWB.
Our results indicate that adding UWB to future motion sensors is worthwhile, though at the cost of requiring calibration across operating temperatures.
UWB is becoming increasingly available as part of location trackers (e.g., Airtags, Tile Ultra, Samsung SmartTag+).

Similar to other learning-based methods, our graph-based pipeline has limits predicting out-of-distribution poses.
Our evaluation assumed a flat terrain and covered simple interaction with objects in the environment.
A more robust method will require a dataset that captures a wider variety of motions and interactions.

Our method marginally improves global translation estimates, with a 0.318\,m and 0.422\,m error for ground-truth root movements of 2\,m and 5\,m, respectively.
This is close to PIP's 0.333\,m and 0.440\,m error.
While distances constrain cross-joint motion, global translation benefits from better pose prediction.
Distance constraint-based translation estimation could be interesting in future work.

Finally, our state and distance estimates from real data are error-prone, whereas our evaluation on synthetic datasets with drift-free orientations and perfect distances shows our method's potential.

\section{Conclusion}
\label{sec:conclusion}

We have proposed a novel method for sparse, wearable sensor-based full-body pose tracking that is independent of visual input.
It leverages a novel source of tracking input: inter-sensor distances estimated from UWB-based ranging, which we use to stabilize the raw signals from IMUs.
We have demonstrated that based on six tracking nodes, our method affords training on existing motion capture datasets with simulated distances to robustly estimate full-body poses.
Our tracking nodes integrate off-the-shelf components found in emerging UWB location trackers, and our pose estimation operates without the need for 3D state estimation from proprietary sensors or tracking systems, making \projname\ a scalable approach for sparse human motion tracking in the wild.


\begin{acks}
We sincerely thank Xintong Liu and Max Moebus for their support and Paul Streli for feedback.
We also thank all the participants of our data capture for their time.
\end{acks}

\balance
\bibliographystyle{ACM-Reference-Format}
\bibliography{main}

\begin{figure*}
    \centering
    \includegraphics[width=0.85\linewidth]{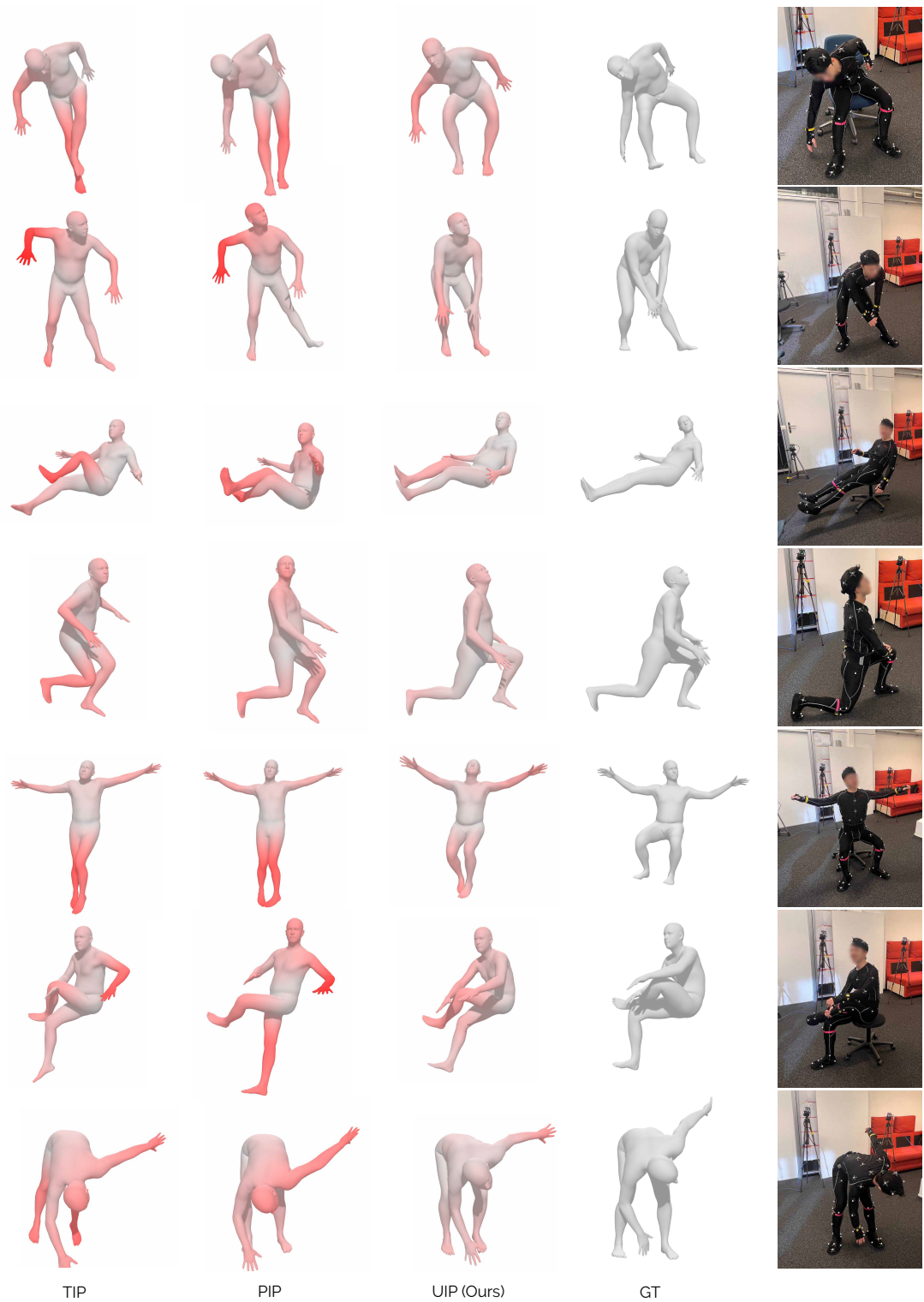}
    \caption{Qualitative comparisons among different methods on our test set.}
    \Description{
    The figure presents a grid of SMPL avatars, for a visual comparison of pose estimation results for different poses and methods. The grid has 5 columns and 7 rows. The left-most column corresponds to pose estimates from TIP, the next to estimates from PIP, then our method UIP followed by ground truth. The right-most column shows photographs of the subject performing the reconstructed poses while wearing a motion capture suit. Each row illustrates a different pose. From the top, the first, third, fifth and sixth rows show sitting poses, while the remaining rows demonstrate various standing poses. The various SMPL reconstructions are grey with red highlights on the limbs, indicating estimating errors. Overall, pose estimates from TIP and PIP exhibit more pronounced errors, especially around the arms and legs.
    }
    \label{fig:compare_long}
\end{figure*}

\clearpage
\end{document}